# LP-DETR: Layer-wise Progressive Relation for Object Detection

Zhengjian Kang[1*], Ye Zhang[2*], Xiaoyu Deng[3], Xintao Li[4], Yongzhe Zhang[5]

[1] New York University, NY 10012, USA
[2] University of Pittsburgh, PA 15213, USA
[3] Fordham University, NY 10458, USA
[4] Georgia Institute of Technology, GA 30332, USA
[5] California Institute of Technology, CA 91125, USA
[1]zk299@nyu.edu, [2]yez12@pitt.edu, [3]xdeng24@fordham.edu, [4]xli3204@gatech.edu, [5]yongzhe@caltech.edu

**Abstract.** This paper presents LP-DETR (Layer-wise Progressive Relation DETR), a novel approach that enhances DETR-based object detection through multi-scale relation modeling. Our method introduces learnable spatial relationships between object queries through a relation-aware self-attention mechanism, which adaptively learns to balance different scales of relations (local, medium, and global) across decoder layers. This progressive design enables the model to effectively capture evolving spatial dependencies throughout the detection pipeline. Extensive experiments on the COCO 2017 dataset demonstrate that our method improves both convergence speed and detection accuracy compared to the standard self-attention module. The proposed method achieves competitive results, reaching 52.3% AP with 12 epochs and 52.5% AP with 24 epochs using the ResNet-50 backbone, and further improving to 58.0% AP with the Swin-L backbone. Furthermore, our analysis reveals a learning pattern: The model naturally learns to prioritize local spatial relations in early decoder layers, while gradually shifting attention to broader contexts in deeper layers, providing valuable insights about spatial relations for future research in object detection.

## 1 Introduction

Object detection, a core computer vision technique, enables automated identification and localization of objects within images and videos. This technology has diverse applications spanning urban planning and surveillance [26], industrial defect detection [27], and bioinformatics-driven drug discovery [19]. Detection Transformers (DETRs) [3] have achieved great progress by proposing an end-to-end architecture in the field of object detection. However, their low training efficacy remains a critical challenge. The root cause is the imbalanced supervision during training - DETR employs a Hungarian algorithm to assign only one positive prediction to each ground-truth box, leaving the majority of predictions as negative samples. This insufficient positive supervision leads to slow and unstable convergence. Although various approaches have been proposed to address this issue through different technical routes such as multi-scale feature learning [37], denoising training [17, 32], hybrid matching strategies [4, 38] and loss alignment [2, 23], they primarily focus on local feature enhancement or individual query optimization, leaving the potential of relation modeling of the object queries in self-attention not fully explored.

---

[*] Equal contribution

In the vision community, modeling inter-object relationships has proven beneficial for detection performance. Previous approaches mainly focus on two aspects: co-occurrence patterns of object categories [5, 8, 16, 30] and spatial relations using various criteria [12, 28, 35]. These methods have demonstrated that incorporating relation information can effectively enhance detection accuracy by capturing contextual dependencies between objects. However, in the DETR field, few works have investigated the learnable relation between object queries in self-attention, a key component in DETR decoders. Hao et al. [8] first attempt to model class correlations using a learnable relation matrix in the decoder self-attention, but their approach does not consider spatial information and requires mapping class-to-class relations back to query-to-query interactions. More recently, Relation-DETR [10] introduces explicit position relations between bounding boxes with cross-layer refinement. Motivated by their work, but different from these approaches, we directly incorporate geometric relation weights into queries within each layer and propose layer-wise progressive relation modeling to capture evolving multi-scale spatial dependencies.

In this paper, we present LP-DETR (Layer-wise Progressive Relation DETR), which enhances object detection through explicit modeling of multi-scale spatial relations across decoder layers. Our key insight is that object relations naturally evolve from local to global contexts through the detection pipeline, and different scales of spatial relations may play varying roles at different stages of the detection process. We propose a progressive relation-aware self-attention module that adaptively learns to balance different scales of spatial relations (local, medium, and global) at different decoder layers.

The main contributions of our work are threefold.

- We introduce a relation-aware self-attention mechanism that explicitly models multi-scale spatial relationships between object queries.
- We propose a progressive refinement strategy that allows the model to adaptively adjust the relation weights across the decoder layers.
- We validate a learning pattern where spatial relations naturally progress from local to global contexts in decoder layers, providing valuable insights for future research.

Extensive experiments on the COCO 2017 dataset demonstrate the effectiveness of our approach. LP-DETR achieves competitive results with 52.3% AP under 12-epoch training and 52.5% AP under 24-epoch training using ResNet-50 backbone. With Swin-L backbone, our method improves further to 58.0% AP. More importantly, our analysis reveals that the proposed progressive relation modeling contributes to both improved convergence and detection accuracy. These results validate our hypothesis on the importance of layer-wise progressive relation modeling and suggest promising directions for future research in object detection.

## 2 Related Work

### 2.1 Transformer for Object Detection

DEtection TRansformer (DETR) [3] establishes a new paradigm for end-to-end object detection by eliminating hand-crafted post-processing steps such as Non-maximum Suppression (NMS). Its transformer-based architecture consists of two main components: an encoder that transforms image features into enriched memory representations and a decoder that converts a set of learnable object queries into detection results. The decoder operates through two attention mechanisms: self-attention for modeling interactions among object queries, and cross-attention for capturing relationships between queries and encoded memories.

However, DETR suffers from slow convergence during training. Various approaches have been proposed to address this issue from different perspectives: (1) **Enhanced Feature Learning**: Deformable DETR [37] explores multi-scale features through deformable attention with sparse reference points, while Focus-DETR [36] and Salience-DETR [11] improve feature selection through salient token identification in the encoder. (2) **Query Enhancement**: DAB-DETR [22] decouples object queries into anchor box for iterative refinement, while DN-DETR [17] and DINO [32] accelerate training through auxiliary denoising task and contrastive learning. (3) **Better Supervision**: Hybrid DETR [38] and Group DETR [4] adopt one-to-many matching to increase supervision signals, while Stable-DINO [23] and Align-DETR [2] propose specialized loss functions to align classification and localization. (4) **Attention Mechanism**: Recent works focus on improving attention mechanism, where Cascade-DETR [31] enhances query-feature interactions through cross-attention, and Relation-DETR [11] learns explicit relation modeling between queries in self-attention.

### 2.2 Relation Network

Relation networks have emerged as a powerful approach for modeling inter-object relationships at instance level, which can be broadly categorized into two main categories: co-occurrence modeling and spatial relation modeling.

Co-occurrence approaches focus on capturing statistical dependencies between object categories. Some methods [16, 30] directly learn from category distribution in large datasets, while others [5, 8] adaptively learn class relationships from annotations. However, these approaches either rely on fixed statistical priors or face challenges in mapping between instances and categories.

Spatial relation approaches, on the other hand, construct graph structures where object features serve as nodes and their spatial relationships as edges. Pioneering works like Relation Network [12] introduces geometric weights in attention modules to model spatial relationships to enhance feature representation. Recent methods determine relation weights through various learned metrics, such as position-aware distance [1, 20], attention mechanisms [28, 35], or appearance similarity [18]. While these learnable relations offer greater flexibility compared to fixed priors, they typically require larger datasets and longer training time to effectively learn the relations from data.

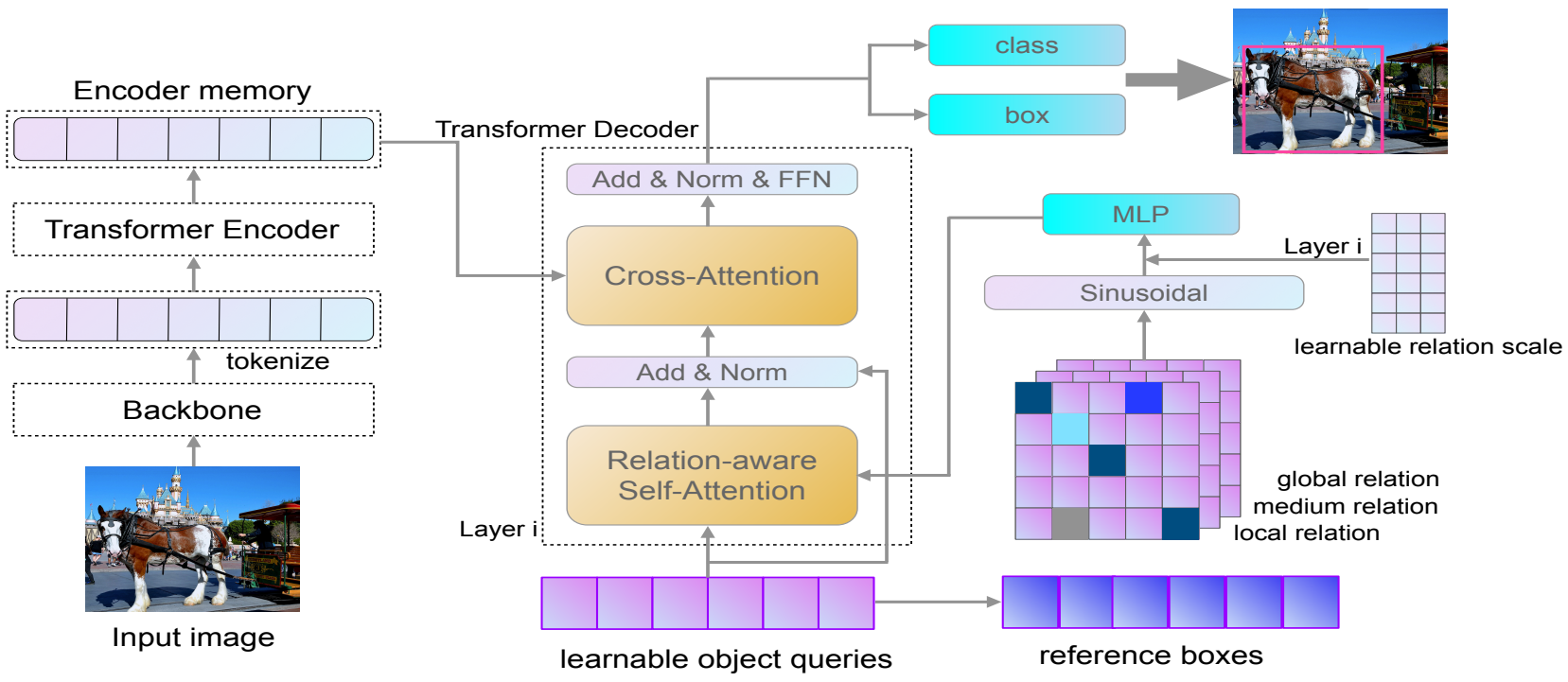

**Fig. 1.** DETR with layer-wise progressive relation pipeline. A learnable relation-aware self-attention mechanism that augments object queries with multi-scale spatial relations, which adaptively evolve from local to global contexts across decoder layers for progressive refinement.

## 3 Methodology

### 3.1 DETR Preliminaries

A DETR-style detector consists of a backbone network (e.g., ResNet [9], Swin Transformer [24]) and a transformer architecture with encoder and decoder modules. Given an input image, the backbone first extracts image features, which are then split into patch tokens. The transformer encoder processes these tokens through self-attention mechanisms and outputs enhanced feature representations, denoted as memories $\mathbf{Z} = \{z_1, \ldots, z_m\}$. The transformer decoder takes as input a set of learnable object queries $\mathbf{Q} = \{q_1, \ldots, q_n\}$. Recent works [22, 32] propose to decouple these queries into content queries $\mathbf{Q}^\mathbf{c}$ for label embedding and position queries $\mathbf{Q}^\mathbf{p}$ for bounding box prediction, allowing better alignment of the relationships. The decoder consists of $L$ stacked blocks, where each block contains three sequential components: a self-attention layer, a cross-attention layer, and a feed-forward network (FFN). The self-attention layer enables communication between object queries, allowing each query to refine its prediction by considering other queries' predictions. The cross-attention layer facilitates interaction between object queries $\mathbf{Q}$ and encoded memories $\mathbf{Z}$. Finally, the FFN transforms the query embeddings for prediction through parallel classification and regression heads.

### 3.2 Layer-wise Progressive Relation-Aware Attention

We briefly introduce the high-level architecture upon transformer-based DETR model in Fig. 1. Our proposed progressive relation-aware attention is applied into the self-attention in the decoder component. Let's consider object queries $\mathbf{Q}$ consists of label embedding queries $\mathbf{Q}^\mathbf{c}$, reference box position queries $\mathbf{Q}^\mathbf{p}$ (represented by $(x, y, w, h)$), and relation queries $\mathbf{Q^r}$. Given a set of $N$ object queries, $q_i = \{q_i^c, q_i^p, q_i^r\}_{i=1}^N$, the $i$-th relation query $q_i^r$ with respect to all the object queries can be calculated as the weighted sum of all the queries:

$$q_i^r = \sum_j^N w_{ij}^r \cdot (W_V \cdot q_j) \tag{1}$$

The relation weight $w_{ij}^r$ captures both geometric and attention-based relationships between queries, which is computed as:

$$w_{ij}^r = \frac{w_{ij}^p \cdot \exp(w_{ij}^a)}{\sum_k^N w_{ik}^p \cdot \exp(w_{ik}^a)}, \tag{2}$$

$$w_{ij}^a = \frac{(W_Q \cdot q_i) \cdot (W_K \cdot q_j)}{\sqrt{d_k}}, \tag{3}$$

where $W_Q$, $W_K$ and $W_V$ are learnable projection matrices for query, key, and value in self-attention, $d_k$ denotes the embedding size of $W_Q \cdot q_i$. The attention scores $w_{ij}^a$ are normalized by geometric weights $w_{ij}^p$ to obtain the final relation weights $w_{ij}^r$. The geometric weight $w_{ij}^p$ incorporates spatial relationships through:

$$w_{ij}^p = W_G \cdot E\left(R(q_i^p, q_j^p)\right), \tag{4}$$

$$R(q_i^p, q_j^p) = \left(\log\left(\frac{|x_i - x_j|}{w_i}\right), \log\left(\frac{|y_i - y_j|}{h_i}\right), \log\left(\frac{w_i}{w_j}\right), \log\left(\frac{h_i}{h_j}\right), \mathrm{IoU}(q_i^p, q_j^p)\right), \tag{5}$$

where the relation metric $R$ captures spatial transformations in distances, scales and IoU. The embedding function $E$ maps these 5-D features to high-dimensional space using sinusoidal encoding [28], followed by a learnable projection $W_G$ implemented as MLP with ReLU activation. The final query representation integrates information from multiple attention heads:

$$q_i = \mathrm{MLP}\left(\mathrm{Concat}(q_i, q_i^{r1}, \dots, q_i^{rK})\right), \tag{6}$$

where $K$ denotes the number of relation-aware attention heads. The Concat operator aggregates the original query and relation queries, and MLP enhances the object queries with learnable query-to-query relation weights, making them more sensitive to spatial relations during training.

To investigate how different scales of relations evolve across decoder layers, we propose three types of relation metrics: local, medium, and global relations. The local relation $R_l$ uses the original metric in Eq. 5, emphasizing relative distances and scale variations. The medium relation $R_m$ applies a scaling factor of $(1 + 2 \times l/L)$ to reduce the log function's steepness. The global relation $R_g$ uses constant weights 1.0. The geometric weight at the $l$-th decoder layer is then formulated as:

$$w_{ij}^{p} = W_G \cdot \Lambda_l \cdot E\left(R_l(q_i^p, q_j^p); R_m(q_i^p, q_j^p); R_g(q_i^p, q_j^p)\right), \quad (7)$$

where $\Lambda = [\lambda_l; \lambda_m; \lambda_g]_{L\times 3}$ represents learnable weights that adaptively adjust the importance of different relation scales across decoder layers.

# 4 Experiments

## 4.1 Experiment Settings

**Dataset and backbone.** We evaluate our layer-wise progressive relation DETR on COCO 2017 [21], CSD [29] and MSSD [6] datasets. COCO 2017 contains 113k training images across 80 object categories. CSD and MSSD are task-specific datasets including 373 training images with 3 categories and 962 training images with 4 categories, respectively. The performance is evaluated using standard COCO metrics: average precision (AP) at different IoU thresholds (IoU=0.5, 0.75, 0.5:0.95) and object scales (small, medium, large). We implement our method with two backbone networks: ResNet50 [9] pretrained on ImageNet-1k and Swin-Large [24] pretrained on ImageNet-22k [7]. Both backbones are finetuned with an initial learning rate of $1 \times 10^{-5}$, which is decreased by a factor of 0.1 at later training stages.
**Implementation details.** All experiments are conducted on NVIDIA RTX 3090 GPUs using AdamW optimizer [15] with a weight decay of $1 \times 10^{-4}$ and a total batch size of 16. For the relation embedding module, we set the temperature $T = 10000$, scale $s = 100$, and position embedding dimension $d_{\text{pos}} = 16$ in sinusoidal encoding. Position relations are constructed at three scales (local, medium, and global) with equal initial weights on 6 decoder layers. Following standard practice in DETR-based methods, we apply common data augmentations including random resize, crop, and flip during training. We report results under both 1x (12 epochs) and 2x (24 epochs) training schedules for COCO 2017, 60 epochs for CSD and 120 epochs for MSSD.

## 4.2 Main Results

**Comparison on COCO 2017.** We evaluate our model on the COCO 2017 validation dataset. The results are summarized in Table 1 and Table 2. Using ResNet-50 backbone with 12-epoch training, our model achieves competitive results of 52.3 AP, 69.6 $\text{AP}_{50}$ and 56.8 $\text{AP}_{75}$. We observe consistent improvements across all metrics (+0.6 AP, +0.5 $\text{AP}_{50}$, +0.5 $\text{AP}_{75}$) compared to Relation-DETR [10]. In terms of different object scales, our method achieves 35.8 $\text{AP}_S$, 55.9 $\text{AP}_M$, and 66.6 $\text{AP}_L$, with notable gains on medium (+0.3 $\text{AP}_M$) and large objects (+0.5 $\text{AP}_L$). These improvements become more evident with 24-epoch training, where our method surpasses Relation-DETR by 0.4 AP, 0.3 $\text{AP}_{50}$ and 0.6 $\text{AP}_{75}$. Our approach also demonstrates strong scalability to larger backbones. With Swin-L backbone and 12-epoch training, we achieve a state-of-the-art performance of 58.1 AP, improving upon the best results by 0.3 AP. We attribute these improvements to our novel relation mechanism, which enables reference boxes to interact with each other in a layer-wise progressive manner.

**Table 1.** Evaluation on COCO 2017 with SOTA methods using ResNet-50.

| Method | Backbone | Epochs | AP | AP_50 | AP_75 | AP_S | AP_M | AP_L |
|---|---|---|---|---|---|---|---|---|
| Def-DETR [37] | ResNet-50 | 50 | 46.2 | 65.2 | 50.0 | 28.8 | 49.2 | 61.7 |
| DAB-DETR [22] | ResNet-50 | 50 | 42.6 | 63.2 | 45.6 | 21.8 | 46.2 | 61.1 |
| DN-Def-DETR [17] | ResNet-50 | 12 | 46.0 | 63.8 | 49.9 | 27.7 | 49.1 | 62.3 |
| DINO [32] | ResNet-50 | 12 | 49.7 | 67.0 | 54.4 | 31.4 | 52.9 | 63.6 |
| H-Def-DETR [14] | ResNet-50 | 12 | 48.7 | 66.4 | 52.9 | 31.2 | 51.5 | 63.5 |
| Cascade-DETR [31] | ResNet-50 | 12 | 49.7 | 67.1 | 54.1 | 32.4 | 53.5 | 65.1 |
| Co-Def-DETR [38] | ResNet-50 | 12 | 49.5 | 67.6 | 54.3 | 32.4 | 52.7 | 63.7 |
| Align-DETR [2] | ResNet-50 | 12 | 50.2 | 67.8 | 54.4 | 32.9 | 53.3 | 65.0 |
| Stable-DINO [23] | ResNet-50 | 12 | 50.4 | 67.4 | 55.0 | 32.9 | 54.0 | 65.5 |
| DAC-DETR [13] | ResNet-50 | 12 | 50.0 | 67.6 | 54.7 | 32.9 | 53.1 | 64.4 |
| Rank-DETR [25] | ResNet-50 | 12 | 50.4 | 67.9 | 55.2 | 33.6 | 53.8 | 64.2 |
| MS-DETR [34] | ResNet-50 | 12 | 50.3 | 67.4 | 55.1 | 32.7 | 54.0 | 64.6 |
| Relation-DETR [10] | ResNet-50 | 12 | 51.7 | 69.1 | 56.3 | **36.1** | 55.6 | 66.1 |
| LP-DETR | ResNet-50 | 12 | **52.3** | **69.6** | **56.8** | 35.8 | **55.9** | **66.6** |
| H-Def-DETR [14] | ResNet-50 | 36 | 50.0 | 68.3 | 54.4 | 32.9 | 52.7 | 65.3 |
| DINO [32] | ResNet-50 | 24 | 50.4 | 68.3 | 54.8 | 33.3 | 53.7 | 64.8 |
| DDQ-DETR [33] | ResNet-50 | 24 | 52.0 | 69.5 | **57.2** | 35.2 | 54.9 | 65.9 |
| Relation-DETR [10] | ResNet-50 | 24 | 52.1 | 69.7 | 56.6 | 36.1 | 56.0 | 66.5 |
| LP-DETR | ResNet-50 | 24 | **52.5** | **70.0** | **57.2** | **36.2** | **56.3** | **67.1** |

**Table 2.** Evaluation on COCO 2017 with SOTA methods using Swin-L.

| Method | Backbone | Epochs | AP | AP_50 | AP_75 | AP_S | AP_M | AP_L |
|---|---|---|---|---|---|---|---|---|
| DINO [32] | Swin-L | 12 | 56.8 | 75.4 | 62.3 | 41.1 | 60.6 | 73.5 |
| H-Def-DETR [14] | Swin-L | 12 | 55.9 | 75.2 | 61.0 | 39.1 | 59.9 | 72.2 |
| Rank-DETR [25] | Swin-L | 12 | 57.3 | 75.9 | 62.9 | 40.8 | 61.3 | 73.2 |
| Relation-DETR [10] | Swin-L | 12 | 57.8 | 76.1 | 62.9 | **41.2** | 62.1 | 74.4 |
| LP-DETR | Swin-L | 12 | **58.1** | **76.4** | **63.2** | 41.0 | **62.2** | **74.7** |

**Comparison on CSD and MSSD.** To further validate the generalization capability of our proposed method, we conduct extensive evaluations on CSD and MSSD datasets. As shown in **Table 3**, LP-DETR outperforms Relation-DETR by 0.4 AP on CSD, achieving state-of-the-art performance with 54.8% AP. Moreover, the results in **Table 4** demonstrate that our method improves the performance to 58.0 AP on MSSD, significantly surpassing all existing approaches.

### 4.3 Ablation study

We conduct a few ablation studies to investigate the performance effect of different components of our model on the COCO dataset.

**Table 3.** Evaluation on CSD with SOTA methods using ResNet-50.

| Method | Backbone | Epochs | AP | AP_50 | AP_75 | AP_S | AP_M | AP_L |
|---|---|---|---|---|---|---|---|---|
| DAB-Def-DETR [22] | ResNet-50 | 90 | 52.9 | 91.2 | 55.0 | 50.3 | 39.4 | 0.0 |
| DN-Def-DETR [17] | ResNet-50 | 60 | 49.9 | 88.0 | 51.2 | 47.6 | 37.7 | 0.0 |
| DINO [32] | ResNet-50 | 60 | 53.0 | 90.8 | 55.5 | 50.9 | 39.6 | 0.0 |
| Relation-DETR [10] | ResNet-50 | 60 | 54.4 | 92.9 | 56.0 | 53.6 | 40.8 | 0.0 |
| LP-DETR | ResNet-50 | 60 | **54.8** | **93.4** | **56.2** | **53.9** | **41.2** | 0.0 |

**Table 4.** Evaluation on MSSD with SOTA methods using ResNet-50.

| Method | Backbone | Epochs | AP | AP_50 | AP_75 | AP_S | AP_M | AP_L |
|---|---|---|---|---|---|---|---|---|
| DAB-Def-DETR [22] | ResNet-50 | 120 | 33.7 | 60.0 | 31.0 | 15.9 | 26.7 | 29.0 |
| DN-Def-DETR [17] | ResNet-50 | 120 | 45.6 | 74.1 | 44.9 | 18.6 | 31.9 | 41.0 |
| DINO [32] | ResNet-50 | 120 | 51.0 | 80.0 | 52.5 | 20.0 | 47.4 | 44.8 |
| Relation-DETR [10] | ResNet-50 | 120 | 57.4 | 79.2 | 63.6 | 31.4 | 53.5 | 45.5 |
| LP-DETR | ResNet-50 | 120 | **58.0** | **80.0** | **64.0** | **31.8** | **54.2** | **46.1** |

**Analysis performance on the number of relation heads.** We examine how the number of relation heads affects model performance in our progressive relation self-attention module. Fig. 2a shows the results with varying numbers of relation heads while maintaining a total of 8 attention heads. Using no relation head (0 head) serves as our baseline, where the module functions as standard self-attention, achieving 51.1 AP. The performance consistently improves as we increase the number of relation heads. With all 8 heads dedicated to relation-aware attention, our model achieves the best performance of 52.3 AP, showing an improvement of 1.2 AP over the baseline. These results demonstrate the benefit of incorporating relation-aware attention in the self-attention mechanism.

**Analysis performance on the different IoUs.** As shown in Fig. 2b, we evaluate the impact of different IoU variants on detection performance by incorporating standard IoU, GIoU, DIoU, and CIoU into our relation metric. We adopt CIoU as our IoU choice as our experimental results indicate that it achieves marginally superior performance compared to other variants.

**Analysis performance on the different modules.** We evaluate the effectiveness of progressive refinement (PR) in combination with our relation-aware self-attention module. As shown in Table 5, the relation-aware self-attention module alone improves the detection performance by 0.9 AP, 0.6 $AP_{50}$, and 0.7 $AP_{75}$ compared to the baseline. Adding progressive refinement brings additional gains of 0.3 AP, 0.4 $AP_{50}$, and 0.3 $AP_{75}$. The improvements are consistent across different object scales, with enhanced performance in $AP_S$, $AP_M$ and $AP_L$.

### 4.4 Progressive Refinement Analysis

To understand how different scales of spatial relations (local, medium, and global) evolve through decoder layers, we visualize the learned relation weights across layers in Fig. 3a. Our analysis reveals an interesting pattern in how the model balances different spatial relations. In the first layer, the weights among the three scales remain relatively comparable, suggesting initial uncertainty in relation to scale selection. From layer 2, the model develops a strong preference for local relations, with weights exceeding 0.9, indicating early decoder layers focus primarily on establishing local relationships between queries. Moving to deeper layers, we observe a gradual transition: local relation weights decrease to 0.24, while medium and global relation weights steadily increase to around 0.4. By the final layer, the weights for medium and global relations surpass that of local relations. This progressive transition from local to broader spatial contexts aligns with the intuitive understanding that object detection requires hierarchical processing - from local to global semantic interpretation.

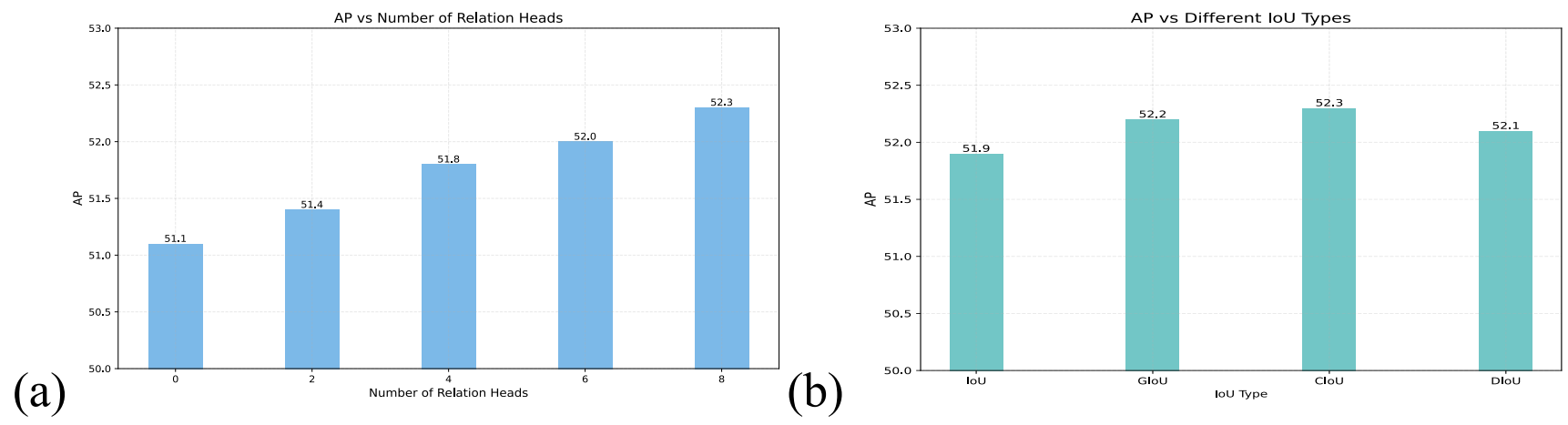

**Fig. 2.** (a) Performance comparison of different numbers of relation heads in Relation-Aware self-attention module. (b) Performance comparison of different IoUs in the relation metric.

**Table 5.** Performance comparison on different components used in the detector.

| Component | AP | AP_50 | AP_75 | AP_S | AP_M | AP_L |
|---|---|---|---|---|---|---|
| | 51.1 | 68.6 | 55.8 | 35.1 | 55.2 | 66.0 |
| Relation | 52.0(**↑0.9**) | 69.2(**↑0.6**) | 56.5(**↑0.7**) | 35.6(**↑0.5**) | 55.5(**↑0.3**) | 66.4(**↑0.4**) |
| Relation+PR | 52.3(**↑0.3**) | 69.6(**↑0.4**) | 56.8(**↑0.3**) | 35.8(**↑0.2**) | 55.9(**↑0.4**) | 66.6(**↑0.2**) |

### 4.5 Convergence comparison

Fig. 3b plots the convergence curves of different state-of-the-art methods with ResNet-50 backbone. Our model shows improved convergence behavior, benefiting from the learned layer-wise multi-scale position relations between object queries. Although the absolute AP gain over Relation-DETR is moderate (+0.6 AP), our method consistently outperforms the baseline throughout the training process. This demonstrates that introducing multi-scale spatial relations can effectively refine the original relation modeling for object detection.

### 4.6 Qualitative Results on Selected Images

To demonstrate the effectiveness of our proposed method, Fig. 4 presents detection results in challenging scenarios with complex spatial relationships. These include crowded traffic scenes, sports fields with multiple human subjects, and various object interactions involving occlusions. Our proposed method improves object detection by leveraging layer-wise progressive relations between object candidates. The model shows particular strength in handling two challenging cases: (1) occluded objects, where partial visibility makes detection difficult using appearance features alone, and (2) densely packed scenes, where spatial and semantic relations between objects provide crucial disambiguation cues.

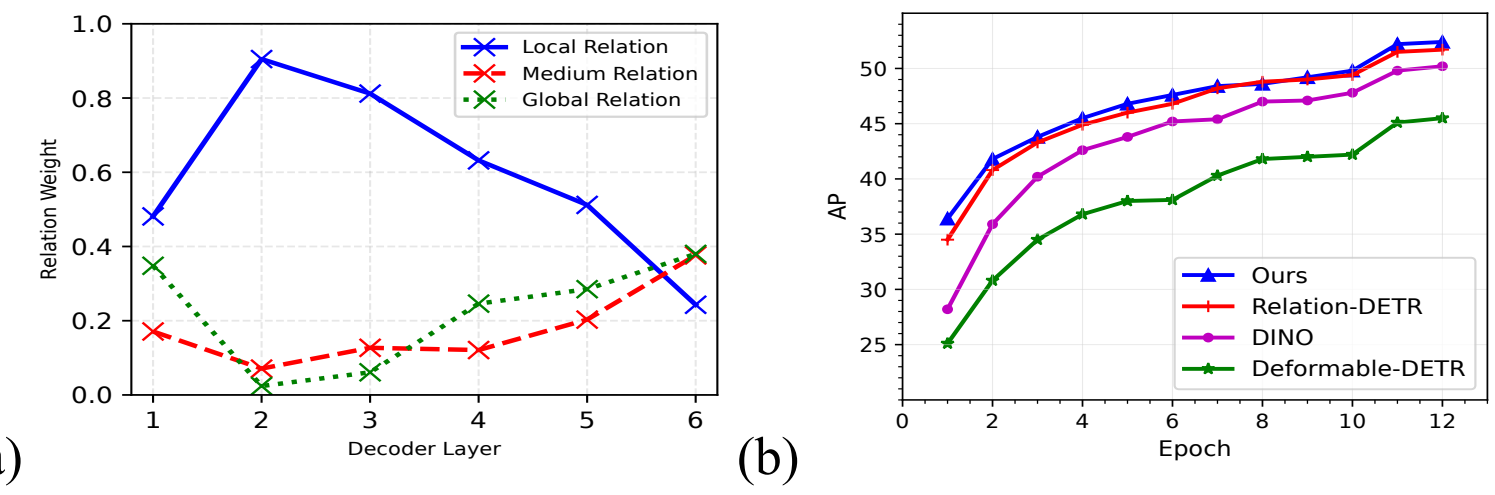

**Fig. 3.** (a) Relation weight under local, medium and global relations in different decoder layers. (b) Convergence comparison under different state-of-the-art methods with ResNet-50 backbone.

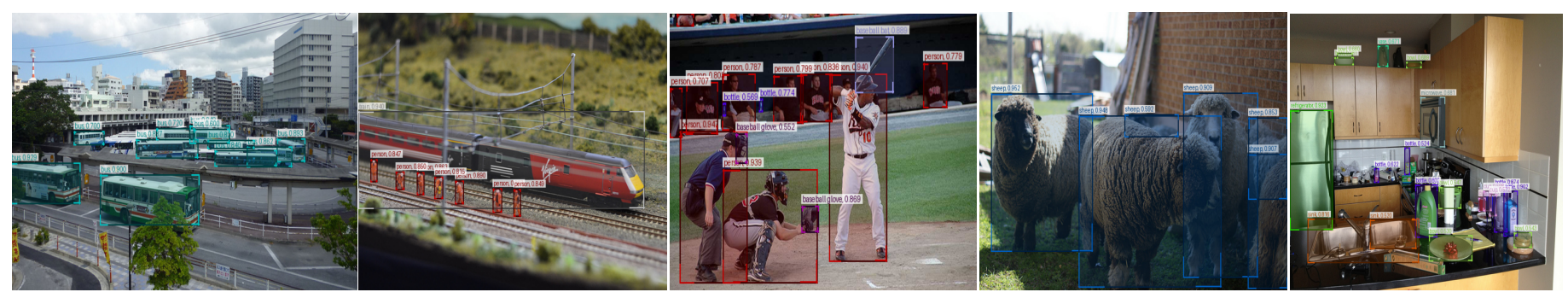

**Fig. 4.** Qualitative detection results in complex scenarios. Our method successfully identifies objects in challenging situations including occlusions, crowded scenes, and complex spatial relationships.

## 5 Conclusion

We present a progressive relation-aware self-attention module that enhances DETR by incorporating learnable multi-scale spatial relationships between object queries. Our approach adaptively adjusts relation weights across scales and decoder layers, improving both convergence speed and detection accuracy on standard benchmarks. Our analysis reveals that local relations dominate in early decoder layers while global relations become increasingly important in deeper layers, aligning with the hierarchical nature of object detection. Despite these advances, our method shows only marginal improvements on small objects, primarily due to the inherent challenges in small object detection — notably the limited number of small objects and their weak feature representations. Future work could investigate learned visual representations for spatial relations, explore how spatial relations can further help improve small object detection, and extend to challenging scenarios with significant occlusions.

**Disclosure of Interests.** The authors have no competing interests to declare that are relevant to the content of this article.